\documentclass{article}

\usepackage{microtype}
\usepackage{graphicx}
\usepackage{booktabs} 

\usepackage{natbib}
\usepackage{amssymb, amsmath,amsthm}
\usepackage{amsfonts}
\usepackage{algorithm}
\usepackage{algorithmic}
\usepackage{paralist}
\usepackage{multirow}
\usepackage{dsfont}
\usepackage{mathrsfs}
\usepackage{times}
\usepackage{ulem}
\usepackage{wrapfig}

\usepackage{url,enumerate}
\usepackage{color,xcolor}
\usepackage{makeidx}  
\usepackage{amsmath,amssymb}
\usepackage{mathtools}
\usepackage[small, compact]{titlesec}
\usepackage{xspace}
\usepackage{epstopdf}
\usepackage{cite}
\usepackage{mathrsfs}
\usepackage{times}
\usepackage{enumerate}
\usepackage{color}
\usepackage{graphicx,epsfig}
\usepackage{amsmath,amssymb,xspace}
\usepackage{url}
\usepackage{hyperref}
\usepackage{bm}
\usepackage{upgreek}
\usepackage{cleveref}
\usepackage{multirow}
\usepackage{ulem}
\usepackage{cancel}
\usepackage{subcaption}

\usepackage{hyperref}

\usepackage[accepted]{icml2022}

\icmltitlerunning{Nonparametric Factor Trajectory Learning for Dynamic Tensor Decomposition}

\newcommand{\ours}{{{NONFAT}}\xspace}

\newcommand{\zsdc}[1]{[\textcolor{blue}{#1}]}
\newcommand{\alanc}[1]{}

\newcommand{\kron}{\otimes}

\newcommand{\wh}[1]{\widehat{#1}}

\renewcommand{\a}{{\bf a}}
\renewcommand{\b}{{\bf b}}

\renewcommand{\d}{{\rm d}}  
\newcommand{\e}{{\bf e}}
\newcommand{\f}{{\bf f}}
\newcommand{\g}{{\bf g}}
\newcommand{\h}{{\bf h}}

\newcommand{\m}{{\bf m}}

\renewcommand{\t}{{\bf t}}
\renewcommand{\u}{{\bf u}}
\renewcommand{\v}{{\bf v}}

\newcommand{\x}{{\bf x}}
\newcommand{\y}{{\bf y}}

\newcommand{\A}{{\bf A}}

\newcommand{\C}{{\bf C}}

\newcommand{\E}{{\bf E}}
\newcommand{\F}{{\bf F}}
\newcommand{\G}{{\bf G}}

\newcommand{\I}{{\bf I}}

\newcommand{\K}{{\bf K}}
\renewcommand{\L}{{\bf L}}

\newcommand{\bell}{{\boldsymbol \ell}}

\newcommand{\whM}{{\widehat{\Mcal}}}
\newcommand{\whf}{{\widehat{f}}}
\newcommand{\N}{\mathcal{N}}  
\newcommand{\MN}{\mathcal{MN}} 
\newcommand{\gp}{\mathcal{GP}} 

\newcommand{\Bcal}{\mathcal{B}}

\newcommand{\Dcal}{\mathcal{D}}
\newcommand{\Ocal}{\mathcal{O}}

\newcommand{\Lcal}{\mathcal{L}}

\newcommand{\R}{{\bf R}}
\renewcommand{\S}{{\bf S}}

\newcommand{\U}{{\bf U}}
\newcommand{\V}{{\bf V}}

\newcommand{\X}{{\bf X}}

\newcommand{\Z}{{\bf Z}}
\newcommand{\Mcal}{{\mathcal{M}}}
\newcommand{\Wcal}{{\mathcal{W}}}
\newcommand{\Ucal}{{\mathcal{U}}}

\newcommand{\balpha}{\boldsymbol{\alpha}}

\newcommand{\blambda}{\boldsymbol{\lambda}}

\newcommand{\bOmega}{\mathbf{\Omega}}
\newcommand{\bomega}{\boldsymbol{\omega}}

\newcommand{\bGamma}{\mathbf{\Gamma}}

\newcommand{\0}{{\bf 0}}

\newcommand{\ben}{\begin{enumerate}}
\newcommand{\een}{\end{enumerate}}

\newcommand{\ie}{{\textit{i.e.,}}\xspace}
\newcommand{\eg}{{\textit{e.g.,}}\xspace}
\newcommand{\etc}{{\textit{etc.}}\xspace}
\newcommand{\EE}{\mathbb{E}}

\newcommand{\cmt}[1]{}

\newcommand{\bi}{{\bf i}}

\newcommand{\kl}{{\mathrm{KL} }}

\begin{document}

\twocolumn[
\icmltitle{Nonparametric Factor Trajectory Learning for Dynamic Tensor Decomposition}

\begin{icmlauthorlist}
\icmlauthor{Zheng Wang}{theU}
\icmlauthor{Shandian Zhe}{theU}
\end{icmlauthorlist}

\icmlaffiliation{theU}{School of Computing, University of Utah}

\icmlcorrespondingauthor{Shandian Zhe}{zhe@cs.utah.edu}

\icmlkeywords{Machine Learning, ICML}
\vskip 0.3in
]
\printAffiliationsAndNotice{}

\begin{abstract}
Tensor decomposition is a fundamental framework to analyze data that can be represented by multi-dimensional arrays. In practice,  tensor data is often accompanied with temporal information, namely the time points when the entry values were generated. This information implies abundant, complex temporal variation patterns. However, current methods always assume the factor representations of the entities in each tensor mode are static, and never consider their temporal evolution. To fill this gap, we propose NONparametric FActor Trajectory learning for dynamic tensor decomposition (\ours).   We place Gaussian process (GP) priors in the frequency domain and conduct inverse Fourier transform via Gauss-Laguerre quadrature to sample the trajectory functions. In this way, we can overcome data sparsity and obtain robust trajectory estimates across long time horizons. Given the trajectory values at specific time points, we use a second-level GP to sample the entry values and to capture the temporal relationship between the entities. For efficient and scalable inference, we leverage the matrix Gaussian structure in the model, 
introduce a matrix Gaussian posterior, and develop a nested sparse variational learning algorithm. We have shown the advantage of our method in several real-world applications. 
\end{abstract}

\section{Introduction}
Data involving interactions between multiple entities can often be represented by multidimensional arrays, \ie tensors, and are ubiquitous in practical applications. For example, a four-mode tensor \textit{(user, advertisement, web-page, device)} can be extracted from logs of an online advertising system, and three-mode \textit{(patient, doctor, drug}) tensor from medical databases. As a powerful approach for tensor data analysis, tensor decomposition estimates a set of latent factors to represent the entities in each mode, and use these factors to reconstruct the observed entry values and to predict missing values.  These factors can be further used to explore hidden structures from the data, \eg via clustering analysis, and provide useful features for important applications, such as personalized recommendation, click-through-rate prediction, and disease diagnosis and treatment.

In practice, tensor data is often accompanied with valuable temporal information, namely the time points at which each interaction took place to generate the entry value. These time points signify that underlying the data can be rich, complex temporal variation patterns. To leverage the temporal information, existing tensor decomposition methods usually introduce a  time mode~\citep{xiong2010temporal,rogers2013multilinear,zhe2016dintucker,zhe2015scalable,du2018probabilistic} and arrange the entries into different time steps, \eg hours or days.  They estimate latent factors for  time steps, and may further model the dynamics between the time factors to better capture the temporal dependencies~\citep{xiong2010temporal}. Recently,  \citet{zhang2021dynamic}  introduced time-varying coefficients in the CANDECOMP/PARAFAC (CP) framework~\citep{Harshman70parafac} to conduct continuous-time decomposition. While successful,  current methods always assume the factors of entities are static and invariant. However, along with the time, these factors, which reflect the entities' hidden properties,  can evolve as well, such as  user preferences, commodity popularity and patient health status. Existing approaches are not able to capture such variations and therefore can miss important temporal patterns.

To overcome this limitation,  we propose \ours, a novel nonparametric  dynamic tensor decomposition model to estimate time-varying factors. Our model is robust, flexible enough to learn various complex trajectories from sparse, noisy data, and capture nonlinear temporal relationships of the entities to predict the entry values. Specifically, we  use Gaussian processes (GPs) to sample frequency functions in the frequency domain, and then generate the factor trajectories via inverse Fourier transform. Due to the nice properties of Fourier bases, we can robustly estimate the factor trajectories across long-term time horizons,  even under sparse and noisy data. We use Gauss-Laguerre quadrature to efficiently compute the inverse Fourier transform. Next, we use a second-level GP to sample entry values at different time points as a function of the corresponding factors values. In this way, we can estimate the complex temporal relationships between the entities.  For efficient and scalable inference, we use the sparse variational GP framework~\citep{GPSVI13} and introduce pseudo inputs and outputs for both levels of GPs. We observe a matrix Gaussian structure in the prior, based on which we can avoid introducing pseudo frequencies, reduce the dimension of pseudo inputs, and hence improve the inference quality and efficiency. We then employ matrix Gaussian posteriors to obtain a tractable variational evidence bound.  Finally, we use a nested reparameterization procedure to implement a stochastic mini-batch variational learning algorithm. 

We evaluated our method in three real-world applications. We compared with the state-of-the-art tensor decomposition methods that  incorporate both continuous and discretized time information.  In most cases, \ours outperforms the competing methods, often by a large margin. \ours also achieves much better test log-likelihood, showing superior posterior inference results. We showcase the learned factor trajectories, which exhibit interesting temporal patterns and extrapolate well to the non-training region. The entry value prediction by \ours also shows a more reasonable uncertainty estimate in both interpolation and extrapolation. 
\section{Preliminaries}\label{sect:bk}
\subsection{Tensor Decomposition}
In general, we denote a $K$-mode tensor or multidimensional array by $\Mcal \in \mathbb{R}^{d_1 \times \ldots \times d_K}$. Each mode consists of $d_k$ entities indexed by $1, \ldots, d_k$. We then denote each tensor entry by $\bell = (\ell_1, \ldots, \ell_K)$, where each element is the entity index in the corresponding mode. The value of the tensor entry, denoted by $m_\bi$,  is the result of interaction between the corresponding entities. For example, given a three-mode tensor \textit{(customer, product, online-store)}, the entry values might be the purchase amount or payment. Given a set of observed entries, tensor decomposition aims to estimate a set of latent factors to represent the entities in each mode. Denote by $\u^k_j$ the factors of  entity $j$ in mode $k$. These factors can reflect hidden properties of the entities, such as customer interest and preference. We denote the collection of the factors in mode $k$ by  $\U^k = [\u^k_1, \ldots, \u^k_{d_k}]^\top$, and by $\Ucal = \{\U^1, \ldots, \U^K\}$ all the factors for the tensor. 

To learn these factors, a tensor decomposition model is used to fit the observed data. For example, Tucker decomposition~\citep{Tucker66} assumes that $\Mcal = \Wcal \times_1 \U^1 \times_2 \ldots \times_K \U^K$, where $\Wcal \in \mathbb{R}^{r_1 \times \ldots \times r_K}$ is called tensor core, and  $\times_k$ is the tensor-matrix product at mode $k$~\citep{kolda2006multilinear}. If we restrict $\Wcal$ to be diagonal, it becomes the popular CANDECOMP/PARAFAC (CP) decomposition~\citep{Harshman70parafac}, where each entry value is decomposed as 
\begin{align}
	m_\bell = (\u^1_{\ell_1} \circ \ldots \circ \u^K_{\ell_K})^\top \blambda, \label{eq:cp}
\end{align}
where $\circ$ is the element-wise multiplication and $\blambda$ corresponds to the diagonal elements of $\Wcal$.  While CP and Tucker are popular and elegant, they assume a multilinear interaction between the entities. In order to flexibly estimate various interactive relationships (\eg from simple linear to highly nonlinear), \citep{xu2012infinite,zhe2015scalable,zhe2016dintucker} view the entry value $m_\bell$ as an unknown function of the latent factors and assign a Gaussian process (GP) prior to jointly learn the function and factors from data, 
\begin{align}
	m_\bell = f(\u^1_{\ell_1}, \ldots, \u^K_{\ell_K}), \;\;\; f \sim \gp\left(0, \kappa(\v_\bell, \v_{\bell'})\right), \label{eq:gptf}
\end{align}
where $\v_\bell = [\u^1_{\ell_1}; \ldots; \u^K_{\ell_K}]$, $\v_{\bell'} = [\u^1_{\bell'_1}; \ldots; \u^K_{\bell'_K}]$ are the factors associated with entry $\bell$ and $\bell'$, respectively, \ie inputs to the function $f$, and $\kappa(\cdot, \cdot)$ is the covariance (kernel) function that characterizes the correlation between function values. For example,  a commonly used one is the square exponential (SE) kernel, $\kappa(\x, \x') = \exp(-\frac{\| \x -\x' \|^2}{\eta})$, where $\eta$ is the kernel parameter. Thus, any finite set of entry values $\m  = [m_{\bi_1}, \ldots, m_{\bi_N}]$, which is a finite projection of the GP, follows a multivariate Gaussian prior distribution, $p(\m) = \N(\m | \0, \K)$ where $\K$ is an $N\times N$ kernel matrix, $[\K]_{n, n'} = \kappa(\v_{\bell_n}, \v_{\bell_{n'}})$.  To fit the observations $\y$, we can  use a noise model $p(\y|\m)$, \eg a Gaussian noisy model for continuous observations.  Given the learned $\m$,  to predict the value of a new entry $\bell^*$,  we can use conditional Gaussian, 
\begin{align}
	p(m_{\bell^*}|\m) = \N(m_{\bell^*}| \mu^*, \nu^*) \label{eq:gp-pred} 
\end{align}
where $\mu^* = \kappa(\v_{\ell^*}, \V) \kappa(\V, \V)^{-1} \m$, $\nu^*=\kappa(\v_{\ell^*}, \v_{\ell^*}) -  \kappa(\v_{\ell^*}, \V)\kappa(\V, \V)^{-1} \kappa(\V, \v_{\ell^*})$, and $\V = [\v_{\bell_1}, \ldots, \v_{\bell_N}]^\top$, because $[\m; m_{\bell^*}]$ also follows a multi-variate Gaussian distribution.

In practice, tensor data is often along with time information, \ie the time point at which each interaction occurred to generate the observed entry value. To exploit this information, current methods partition the time domain into steps $1, 2, \ldots, T$ according to a given interval, \eg one week or month. The observed entries are then binned into the $T$ steps. In this way, a time mode is appended to the original tensor~\citep{xiong2010temporal,rogers2013multilinear,zhe2016dintucker,zhe2015scalable,du2018probabilistic}, $\whM \in \mathbb{R}^{d_1 \times \ldots \times d_K \times T}$.  Then any tensor decomposition method can therefore be applied to estimate the latent factors for both the entities and time steps. To better grasp temporal dependencies, a dynamic model can be used to model the transition between the time factors. For example, \citep{xiong2010temporal} placed a conditional prior over successive steps, $p(\t_{j+1}|\t_j) = N(\t_{j+1}|\t_j, \sigma^2\I)$ where $\t_j$ are the latent factors for time step $j$. 
To leverage continuous time information, the most recent work~\citep{zhang2021dynamic} uses polynomial splines to model the coefficients $\blambda$ in CP decomposition (see \eqref{eq:cp}) as a time-varying (trend) function. 

\subsection{Fourier Transform}
Fourier transform (FT) is a mathematical transform that reveals the connection between functions in the time and frequency domains. In general, for any (complex) integrable function $f(t)$ in the time domain, we can find a corresponding function $\whf(\omega)$ in the frequency domain such that
\begin{align}
	f(t) = \frac{1}{2\pi} \int_{-\infty}^\infty \widehat{f}(\omega) e^{i \omega t} \d \omega,  \label{eq:ift}
\end{align}
where $e^{i \omega t} = \cos(\omega t) + i \sin(\omega t)$, and $i$ indicates the imaginary part. The frequency function can be obtained via a convolution in the time domain,   
\begin{align}
	\whf(\omega) = \int_{-\infty}^\infty f(t) e^{-i \omega t} \d t. \label{eq:fft}
\end{align}
The two functions $(f(t), \whf(\omega))$ is called a Fourier pair. 
Using the time function $f(t)$ to compute the frequency function $\whf(\omega)$, \ie \eqref{eq:fft}, is called forward transform, while using $\whf(\omega)$ to recover $f(t)$, \ie \eqref{eq:ift}, is called inverse transform.  
\cmt{
 technique that transforms a function of time $f(t)$ into a function of frequency $\widehat{f}(\omega)$, which is widely applied in many scientific domains, \eg physics, digital signal processing, and cryptography. The forward FT, \ie analysis equation, breaks a waveform (a function or a signal) into an alternative representation, characterized by sine and cosine functions of varying frequency, according to 
\begin{align*}
	\widehat{f}(\omega) = \int_{-\infty}^\infty f(t) e^{-i \omega t} \d t.
\end{align*}
On the other hand, the inverse FT, \ie synthesis equation, synthesizes the original function from its frequency domain representation, which is
\begin{align*}
	f(t) = \frac{1}{2\pi} \int_{-\infty}^\infty \widehat{f}(\omega) e^{i \omega t} \d \omega.
\end{align*}
}
\begin{figure*}[h]
	\centering
	\setlength\tabcolsep{0pt}
	\captionsetup[subfigure]{aboveskip=0pt,belowskip=0pt}
	\begin{tabular}[c]{c}
		\begin{subfigure}[t]{\textwidth}
			\centering
			\includegraphics[width=\textwidth]{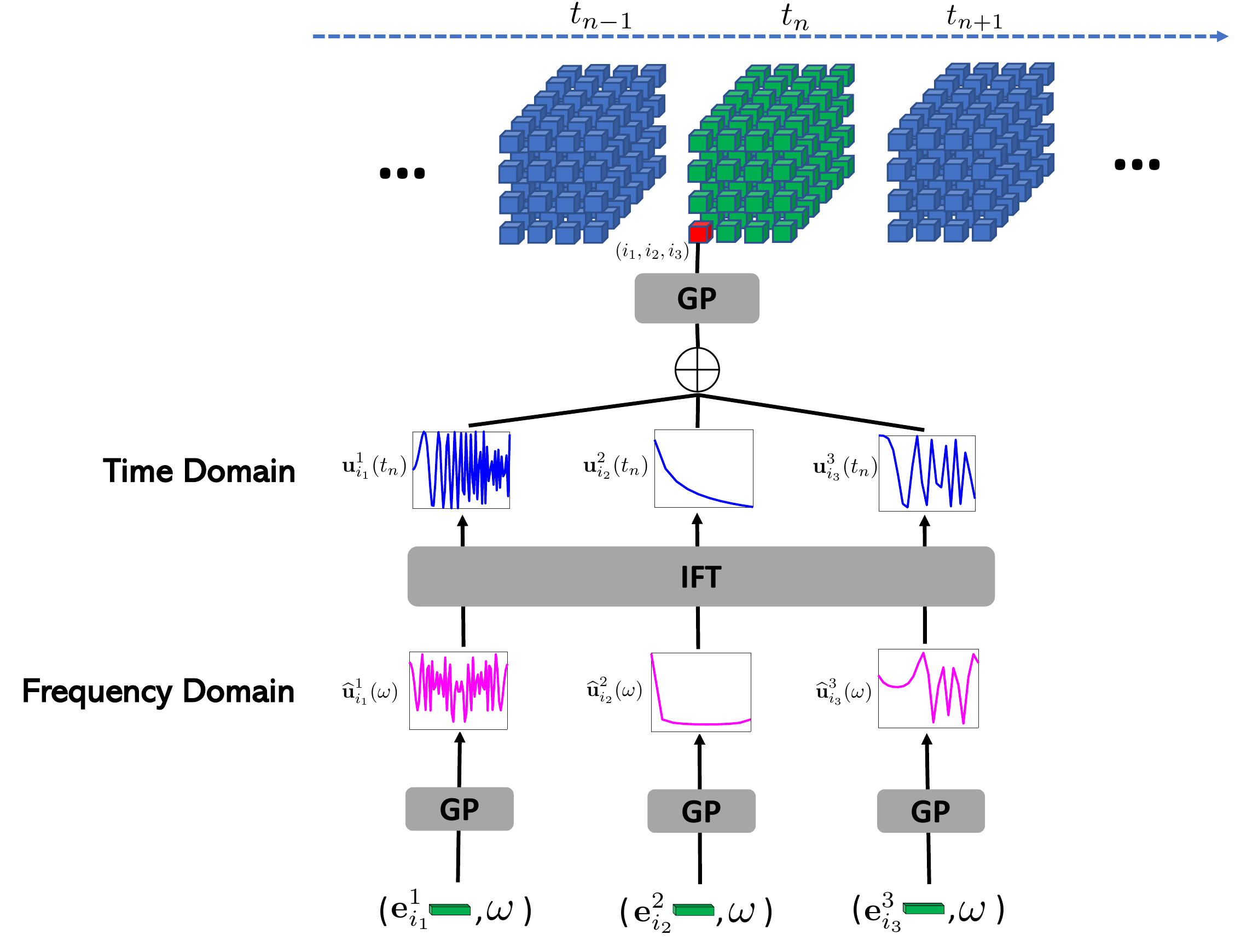}
		\end{subfigure} 
	\end{tabular}
	\caption{\small  A graphical representation of our non-parametric factor trajectory learning model for dynamic tensor decomposition.}
	\label{fig:graphical_model}
\end{figure*}
\section{Model}
Although the existing tensor decomposition methods are useful and successful, they always assume the  factors of the entities are static and invariant, even when the time information is incorporated into the decomposition. However, due to the complexity and diversity of real-world applications, those factors, which reflect the underlying properties of the entities, can evolve over time as well,  \eg customer interest and preference, product quality,  and song popularity. Hence, assuming fixed factors can miss important temporal variation patterns and hinder knowledge discovery and/or downstream predictive tasks. To overcome this issue, we propose \ours, a novel Bayesian nonparametric factor trajectory model for dynamic tensor decomposition.  

Specifically, given the observed tensor entries and their timestamps, $\Dcal = \{(\bell_1, y_1, t_1), \ldots, (\bell_N, y_N, t_N)\}$, we want to learn $R$ factor trajectories (\ie time functions) for each entity $j$ in mode $k$, 
\[
\u^k_j(t) = [u^k_{j,1}(t), \ldots, u^k_{j,R}(t)] \in \mathbb{R}^R.
\]
To flexibly estimate these trajectories, an intuitive idea is to use GPs to model  each $u^k_{j,r}(t)$, similar to that \citet{xu2012infinite,zhe2016dintucker} used GPs to estimate the decomposition function (see \eqref{eq:gptf}).  However, this modeling can be restrictive. When the time point is distant from the training time points, the corresponding covariance (kernel)  function decreases very fast (consider the SE kernel for an example, $\kappa(t, t') = \exp(-\frac{1}{\eta}(t-t')^2)$).  As a result, the GP estimate of the trajectory value --- which is essentially an interpolation based on the training points (see \eqref{eq:gp-pred}) ---  tends to be close to zero (the prior mean) and the predictive variance becomes large.  That means, the GP estimate is neither accurate nor reliable. However, real-world tensor data are often very sparse and noisy --- most entries only have observations at a few time points. Therefore, learning a GP trajectory outright on the time domain might not be robust and reliable enough, especially for many places that are far from the scarce training timestamps.

To address this issue, we find that from the Fourier transform view, any time function can be represented (or decomposed) by a series of Fourier bases $\{e^{i \omega t}|\omega \in \mathbb{R}\}$; see \eqref{eq:ift} and \eqref{eq:fft}.  These bases are trigonometric and have nice properties. We never need to worry that their values will fade to zero (or other constant) when $t$ is large or getting away from the training timestamps. If we can obtain a reasonable estimate of their coefficients,  \ie $\whf(\omega)$ , we can use these bases to recover the time function in a much more reliable way. 

Therefore, we turn to learning the factor trajectories from the frequency domain. Specifically, for each entity $j$ of mode $k$,  we learn a frequency function $\wh{u}^k_{j,r}(\omega)$ ($1 \le r \le R$), so that we can obtain the time trajectory via inverse Fourier transform (IFT),
\begin{align}
	u^k_{j,r}(t) = \frac{1}{2\pi}\int_{-\infty}^{\infty}  \wh{u}^k_{j,r}(\omega) e^{i \omega t} \d \omega.
\end{align}
To get rid of the imaginary part, we require that $\wh{u}^k_{j,r}(\omega)$ is symmetric,  \ie $\wh{u}^k_{j,r}(\omega) = \wh{u}^k_{j,r}(-\omega)$, and therefore we have 
\begin{align}
	u^k_{j,r}(t) = \frac{1}{\pi} \int_{0}^{\infty} \wh{u}^k_{j,r}(\omega) \cos(\omega t)\d \omega. \label{eq:int} 
\end{align}
However, even if the frequency function is given, the integral in \eqref{eq:int} is in general analytically intractable. To overcome this issue, we use Gauss-Laguerre (GL) quadrature, which can solve the integral of the kind  $\int_0^\infty e^{-x}g(x)\d x$ quite accurately.  We write down \eqref{eq:int} as  
\begin{align}
	u^k_{j,r}(t) = \frac{1}{\pi} \int_{0}^{\infty} e^{-\omega}\left[\wh{u}^k_{j,r}(\omega)e^{\omega}\right] \cos(\omega t)\d \omega, 
\end{align}
and then use GPs to learn $\alpha^k_{j,r}(\omega)=\wh{u}^k_{j,r}(\omega)e^{\omega}$. In doing so, not only do we still enjoy the flexibility of nonparametric estimation, we can also conveniently apply GL quadrature, without the need for any additional integral transform, 
 \begin{align}
 	u^k_{j,r}(t) \approx \frac{1}{\pi} \sum_{c=1}^C \alpha^k_{j,r}(\wh{\omega}_c) \cos(\wh{\omega}_c t) \cdot \gamma_c \label{eq:gl}
 \end{align}
where $\{\wh{\omega}_c\}$ and  $\{\gamma_c\}$ are $C$ quadrature nodes and weights.

Next, we introduce a frequency embedding $\e^k_j \in \mathbb{R}^s$ for each entity $j$ in mode $k$, and model $\alpha^k_{j,r}(\cdot)$ as a function of both the embedding and frequency, 
\begin{align}
\alpha^k_{j,r}(\omega) =  f^k_r(\e^k_j, \omega).  
\end{align}
The advantage of doing so is that we only need to estimate one function $f^k_r(\cdot, \cdot)$ to obtain the $r$-th frequency functions for all the entities in mode $k$. The frequency embeddings can also encode structural information within these entities (\eg groups and outliers). Otherwise, we have to estimate $d_k R$ functions in mode $k$, which can be quite costly and challenging, especially when $d_k$ is large.  We then apply a GP prior over $f^k_r$, 
\begin{align}
	 f^k_r(\e, \omega)  \sim\gp\left(0, \kappa_r([\e; \omega], [\e'; \omega']) \right). \label{eq:gp-level1}
\end{align}

Given the factor trajectories, to obtain the value of each entry $m_\bell$ at any time $t$, we use a second-level GP, 
\begin{align}
	&m_{\bell}(t) = g(\u^1_{\ell_1}(t), \ldots, \u^K_{\ell_K}(t)) \notag\\
	& \sim \mathcal{GP}\left(0, \kappa_g(\v_\bell(t), \v_{\bell'}(t))\right), \label{eq:gp-level2}
\end{align}
where $\v_\bell(t) = [\u^1_{\ell_1}(t); \ldots; \u^K_{\ell_K}(t)]$ and $\v_{\bell'}(t) = [\u^1_{\bell'_1}(t); \ldots; \u^K_{\bell'_K}(t)]$. This is similar to \eqref{eq:gptf}. However, since the input consist of the values at (time-varying) trajectories, our second-level GP can flexibly estimate various temporal relationships between the entities. Finally, we sample the observed entry values from a Gaussian noisy model, 
\begin{align}
	p(\y|\m) =  \N(\y | \m, \sigma^2 \I), \label{eq:ll}
\end{align}
where $\sigma^2$ is the noise variance, $\y = \{y_1, \ldots, y_N\}$ and $\m = \{m_{\bell_1}(t_1), \ldots, m_{\bell_N}(t_N)\}$. 
 In this paper, we focus on continuous observations. However, our method  can be easily adjusted to other types of observations.  A graphical illustration of our model is given in Fig. \ref{fig:graphical_model}. 

\section{Algorithm}
The inference of our model is challenging. The GP prior over each $f^k_r(\cdot )$ (see \eqref{eq:gp-level1}) demands we compute a multivariate Gaussian distribution of \cmt{of the function values at all frequency embedding and quadrature nodes combinations,} $\{f^k_r(\e^k_j, \wh{\omega}_c) | 1 \le j \le d_k, 1 \le c \le C\}$, and the GP over $g$ (see \eqref{eq:gp-level2} and \eqref{eq:ll})  a multivariate Gaussian distribution of $\{m_{\bell_n}(t_n)|1\le n\le N\}$.  Hence, when the mode dimensions $d_k$ and/or the number of observations $N$ is large, the computation is very costly or even infeasible, not to mention the two-level GPs are tightly coupled (see \eqref{eq:gp-level2}). To overcome the computational challenges, based on the variational sparse GP framework~\citep{hensman2013gaussian}, we leverage our model structure to develop a nested stochastic  mini-batch variational learning algorithm, presented as follows.

\subsection{Sparse Variational ELBO Based on Matrix Gaussian Prior and Posterior}
Specifically, given $\Dcal = \{(\bell_1, y_1, t_1), \ldots, (\bell_N, y_N, t_N)\}$, the joint probability of our model is
\begin{align}
	p(\text{Joint}) &= \prod_{k=1}^K \prod_{r=1}^R \N\left(\f^k_r|\0, \kappa_r(\X^k_f, \X^k_f)\right) \notag \\
&\cdot  \N\left(\m|\0, \kappa_g(\X_g, \X_g)\right)  \N(\y|\m, \sigma^2\I) \label{eq:joint}
\end{align}
where $\f^k_r$ is the concatenation of  $\{f^k_r(\e^k_j, \wh{\omega}_c)|1 \le j \le d_k, 1\le c \le C\}$, $\m = [m_{\bell_1}(t_1), \ldots, m_{\bell_N}(t_N)]$, $\kappa_r$ and $\kappa_g$ are kernel functions, $\X^k_f$ are $d_k C \times (s+1)$ input matrix for $\f^k_r$, each row of $\X^k_f$ is an $(\e^k_j, \wh{\omega}_c)$ pair,  $\X_g = [\v_{\bell_1}(t_1), \ldots, \v_{\bell_N}(t_N)]^\top$ is the input matrix for $\m$, of size $N \times KR$. In our work, both $\kappa_r$ and $\kappa_r$ are chosen as SE kernels. Note that $\{\wh{\omega}_c\}$ are the quadrature nodes. 

First, we observe that for the first-level GP, the input matrix $\X^k_r$ is  the cross combination of the $d_k$ frequency embeddings $\E^k = [\e^k_1, \ldots, \e^k_{d_k}]^\top$ and $C$ quadrature nodes $\wh{\bomega} = [\wh{\omega}_1; \ldots; \wh{\omega}_C]$. Hence, we can rearrange $\f^k_r$ into a $d_k \times C$ matrix $\F^k_r$. Due to  the multiplicative property of the kernel, \ie $\kappa_r([\e; \omega], [\e'; \omega']) = \kappa_r(\e, \e') \cdot \kappa_r(\omega, \omega') $, we observe $\F^k_r$ follows a matrix Gaussian prior distribution, 
\begin{align}
	&p(\F^k_r) =N(\f^k_r | \0, \kappa_r(\E^k, \E^k) \kron \kappa_r(\wh{\bomega}, \wh{\bomega}) )\notag \\
	&= \MN(\F^k_r|\0,  \kappa_r(\E^k, \E^k), \kappa_r(\wh{\bomega}, \wh{\bomega}))  \label{eq:mg-prior}\\
	&=\frac{\exp\left(-\frac{1}{2}\text{tr}\left( \kappa_r(\wh{\bomega}, \wh{\bomega})^{-1} \left(\F^k_r\right)^\top \kappa_r(\E^k, \E^k)^{-1} \F^k_r\right)\right)}{(2\pi)^{d_k C/2} |\kappa_r(\E^k, \E^k)|^{C/2} |\kappa_r(\wh{\bomega}, \wh{\bomega})|^{d_k/2}}. \notag
\end{align}
Therefore, we can compute the $d_k \times d_k$  row covariance matrix $\kappa_r(\E^k, \E^k)$ and $C \times C$ column covariance matrix $\kappa_r(\wh{\bomega}, \wh{\bomega})$ separately, rather than a giant full covariance matrix, \ie $\kappa_r(\X_f^k, \X_f^k)$ in \eqref{eq:joint} ($d_k C \times d_k C$). This can reduce the cost of the required covariance inverse operation from $\Ocal\left((d_k C)^3\right)$ to $\Ocal(C^3 + (d_k)^3)$ and hence save much cost. Nonetheless, when $d_k$ is large, the row covariance matrix can still be infeasible to compute. Note that  we do not need to worry about the column covariance, because the number of quadrature nodes $C$ is small, \eg $C=10$ is usually sufficient to give a high accuracy in numerical integration.  

To address the challenge of computing the row covariance matrix, we introduce $a_k$ pseudo inputs $\Z^k \in \mathbb{R}^{a_k \times s}$, where $a_k \ll d_k$. We consider the value of $f^k_r(\cdot)$ at the cross combination of $\Z_k$ and $\wh{\bomega}$ , which we refer to as the pseudo output matrix $\G^k_r$.  Then $\{\G^k_r, \F^k_r\}$  follow another matrix Gaussian prior  and we decompose the prior via 
\begin{align}
	p(\G^k_r, \F^k_r) = p(\G^k_r) p(\F^k_r|\G^k_r), \label{eq:joint-prior-level1}
\end{align}
where 
\begin{align}
&p(\G^k_r) = \MN\left(\G^k_r|\0, \kappa_r(\Z^k, \Z^k), \kappa_r(\wh{\bomega}, \wh{\bomega})\right), \notag \\
&p(\F^k_r|\G^k_r) = \MN\left(\F^k_r|\bGamma^k_r, \bOmega^k_r,  \kappa_r(\wh{\bomega}, \wh{\bomega})\right), \label{eq:cmg}
\end{align}
 in which $\bGamma^k_r = \kappa_r(\E^k,\Z^k)\kappa_r(\Z^k, \Z^k)^{-1} \G^k_r$, and $\bOmega^k_r = \kappa_r(\E^k, \E^k) - \kappa_r(\E^k, \Z^k)\kappa_r(\Z^k, \Z^k)^{-1} \kappa_r(\Z^k, \E^k)$.
 
 Similarly, to address the computational challenge for the second-level GP, namely  $\kappa_g(\X_g, \X_g)$ in \eqref{eq:joint}, we introduce another set of $a_g$ pseudo inputs $\Z_g \in \mathbb{R}^{a_g \times KR}$, where $a_g \ll N$.  Denote by $\h$ the corresponding pseudo outputs --- the outputs of $g(\cdot)$ function at $\Z_g$.  We know $\{\m, \g\}$ follow a joint Gaussian prior, and we decompose their prior as well, 
 \begin{align}
 	p(\h, \m) = p(\h)p(\m|\h), \label{eq:joint-prior-level2}
 \end{align}
where $p(\h) = \N\left(\h|\0, \kappa_g(\Z_g, \Z_g)\right)$ and 
\cmt{$p(\m|\h)$ is a conditional Gaussian distribution.} $p(\m|\h) = \N\big(\m|\kappa_g(\X_g,\Z_g)\kappa_g(\Z_g, \Z_g)^{-1}\h, \;\;\kappa_g(\X_g, \X_g) - \kappa_g(\X_g, \Z_g)\kappa_g(\Z_g, \Z_g)^{-1}\kappa_g(\Z_g, \X_g)\big)$.

Now, we augment our model with the pseudo outputs $\{\G^k_r\}$ and $\h$. According to \eqref{eq:joint-prior-level1} and \eqref{eq:joint-prior-level2}, the joint probability becomes
\begin{align}
	&p(\{\G^k_r, \F^k_r\}, \h, \m, \y) = \prod_{k=1}^K \prod_{r=1}^R p(\G^k_r) p(\F^k_r|\G^k_r) \notag \\
	&\cdot  p(\h)p(\m|\h)  \N(\y|\m, \sigma^2\I). \label{eq:joint-2}
\end{align}
Note that if we marginalize out all the pseudo outputs, we recover the original distribution \eqref{eq:joint}.
To conduct tractable and scalable inference, we follow~\citep{hensman2013gaussian} to introduce a variational posterior of the following form, 
\begin{align}
	&q(\{\G^k_r, \F^k_r\}, \h, \m) \notag \\
	&=\prod\nolimits_{k=1}^K\prod\nolimits_{r=1}^R q(\G^k_r) p(\F^k_r |\G^k_r)  q(\h) p(\m|\h). \label{eq:var-post}
\end{align}
We then construct a variational evidence lower bound (ELBO)~\citep{wainwright2008graphical}, $\Lcal = \EE_q\left[\log \frac{p(\{\G^k_r, \F^k_r\}, \h, \m, \y)}{q(\{\G^k_r, \F^k_r\}, \h, \m)}\right]$. Contrasting \eqref{eq:joint-2} and \eqref{eq:var-post}, we can see that all the conditional priors, $p(\F^k_r |\G^k_r)$ and $p(\m|\h)$, which are all giant Gaussian, have been canceled. Then we can obtain a tractable ELBO, which  is additive over the observed entry values, 
\begin{align}
	\Lcal =& -\sum\nolimits_{k=1}^K\sum\nolimits_{r=1}^R\kl\left(q(\G^k_r) \| p(\G^k_r)\right)   \label{eq:elbo} \\
	&- \kl\left(q(\h) \| p(\h)\right)+ \sum\nolimits_{n=1}^N \EE_q\left[\log p(y_n |m_{\bell_n}(t_n))\right], \notag 
\end{align} 
where $\kl(\cdot\| \cdot)$ is the Kullback-Leibler (KL) divergence. We then introduce Gaussian posteriors for $q(\h)$ and all $q(\G^k_r)$ so that the KL terms have closed forms.  Since the number of pseudo inputs $a_g$ and $a_k$ are small (\eg $100$), the cost is cheap.  To further improve the efficiency, we use a matrix Gaussian  posterior for each $\G^k_r$, namely 
\begin{align}
q(\G^k_r) = \MN(\G^k_r | \A^k_r,  \L^k_r\left(\L^k_r\right)^\top, \R^k_r \left(\R^k_r\right)^\top), \label{eq:mgpost}
\end{align}
where $\L^k_r$ and $\R^k_r$ are lower triangular matrices to ensure the row and column covariance matrices are positive semi-definite. Thereby, we do not need to compute the  $a_k C \times a_k C$  full posterior covariance matrix. 
 
 Note that the matrix Gaussian (MG) view \eqref{eq:mg-prior} not only can help us reduce the computation expense, but also improves the approximation. The standard sparse GP approximation requires us to place pseudo inputs in the entire input space. That is, the embeddings plus frequencies.  Our MG view allows us to only introduce pseudo inputs in the embedding space (no need for pseudo frequencies). Hence it decreases approximation dimension and improves inference quality. 

\subsection{Nested Stochastic Mini-Batch Optimization}
We maximize the ELBO \eqref{eq:elbo} to estimate the variational posterior $q(\cdot)$, the frequency embeddings, kernel parameters, pseudo inputs $\Z^k$ and $\Z_g$ and other parameters.\cmt{, so as to obtain the frequency functions $\{f^k_r(\cdot)\}$ and then factor trajectories $\{\u^k_j(t)\}$ (see \eqref{eq:gl}).} To scale to a large number of observations, each step we use a random mini-batch of data points to compute a stochastic gradient, which is based on an unbiased estimate of the ELBO, 
\begin{align}
	\widehat{\Lcal} = \text{KL-terms} + \frac{N}{B} \sum_{n \in \Bcal}\EE_q\left[\log p(y_n |m_{\bell_n}(t_n))\right],
\end{align}
where $\Bcal$ is the mini-batch, of size $B$. However, we cannot  directly compute $\nabla \wh{\Lcal}$ because  each expectation $\EE_q\left[\log p(y_n |m_{\ell_n}(t_n))\right]$ is analytically intractable.  To address this problem, we use a nested reparameterization procedure to compute an unbiased estimate of $\nabla \wh{\Lcal}$, which is also an unbiased estimate of $\nabla \Lcal$, to update the model.
 
 Specifically, we aim to obtain a parameterized sample of each $m_{\bell_n}(t_n)$, plug it in the corresponding log likelihood and get rid of the expectation. Then we calculate the gradient. It guarantees to be an unbiased estimate of $\nabla \wh{\Lcal}$. To do so, following \eqref{eq:joint-prior-level2}, we first draw a parameterized sample of the pseudo output $\h$ via its Gaussian posterior $q(\h)$~\citep{kingma2013auto}, and then apply the conditional Gaussian $p(m_{\bell_n}(t_n)|\h)$.  We denote by $\wh{\h}$ the sample of $\h$. However, the input to $\m_{\ell_n}(t_n)$ are values of the factor trajectories (see \eqref{eq:gp-level2}),  which are modeled by the GPs in the first level. Hence, we need to use the reparameterization trick again to generate a parameterized sample of the input $\v_{\bell_n}(t_n) = [\u^1_{{\ell_n}_1}(t_n); \ldots; \u^K_{{\ell_n}_K}(t_n)]$. To do so, we generate a posterior sample of the pseudo outputs $\{\G^k_r\}$ at the first level. According to \eqref{eq:mgpost}, we can draw a matrix Gaussian noise $\S^k_r \sim \MN(\0, \I, \I)$ and obtain the sample by  $$\wh{\G}^k_r = \A^k_r + \L^k_r  \S^k_r (\R^k_r)^\top.$$ Then we use the conditional matrix Gaussian in \eqref{eq:cmg} to sample $\balpha^k_r \overset{\Delta}{=} [f^k_r(\e^k_{\ell_{n_k}}, \wh{\omega}_1), \ldots, f^k_r(\e^k_{\ell_{n_k}}, \wh{\omega}_C)]$($1 \le k \le K$). Since $\balpha^k_r$ is actually a row of $\F^k_r$,  the conditional matrix Gaussian is degenerated to an ordinary multivariate Gaussian and we generate the parameterized sample $\wh{\balpha}^k_r$ accordingly given $\G^k_r = \wh{\G}^k_r$.  Then we apply the GL quadrature \eqref{eq:gl} to obtain the sample of each trajectory value $\wh{u}^k_{\ell_{n_k},r}(t_n)$, and the input to the second-level GP, $\wh{\v}_{\bell_n}(t_n) = [\wh{\u}^1_{{\ell_n}_1}(t_n); \ldots; \wh{\u}^K_{{\ell_n}_K}(t_n)]$. Now, with the sample of the pseudo output $\wh{\h}$, we can apply $p(m_{\bell_n}(t_n)|\h=\wh{\h})$ to generate the sample of $\m_{\ell_n}(t_n)$. We can use any automatic differentiation library to track the computation of these samples, build computational graphs, and calculate the stochastic gradient.

\subsection{Algorithm Complexity}
The time complexity of our inference algorithm is $\Ocal\left(RC^3+KRa_k^3 + a_g^3 + B(KR(a_k^2 + C^2) + a_g^2) \right)$, in which $B$ is the size of the mini-batch and the cubic terms arise from computing the KL divergence and inverse covariance (kernel) matrix on the pseudo inputs and quadrature nodes (across $K$ modes and $R$ trajectories).  Hence, the cost is proportional to the mini-batch size. The space complexity is $\Ocal(\sum_{k=1}^K\sum_{r=1}^R (a_k^2 + C^2 + a_k C) + a_g^2 + a_g + \sum_{k=1}^K d_k s)$, including the storage of the prior and variational posterior of the pseudo outputs and frequency embeddings. 


\section{Related Work}

 To exploit time information, existent tensor decomposition methods usually expand the tensor with an additional time mode, \eg~\citep{xiong2010temporal,rogers2013multilinear,zhe2016distributed,ahn2021time,zhe2015scalable,du2018probabilistic}. This mode consists of a series of time steps, say by hours or weeks.  A dynamic model is often used to model the transition between the time step factors. For example,  \citet{xiong2010temporal} used a conditional Gaussian prior, \citet{wu2019neural} used recurrent neural networks  to model the time factor transition, and \citet{ahn2021time} proposed a kernel smoothing and regularization term. To deal with continuous time information, the most recent work \citep{zhang2021dynamic} uses polynomial splines to model the CP coefficients ($\blambda$ in \eqref{eq:cp}) as a time function.  Another  line of research focuses on the events of interactions between the entities, \eg  Poisson tensor factorization~\citep{schein2015bayesian,Schein:2016:BPT:3045390.3045686}, and those based on more expressive point processes~\citep{zhe2018stochastic,pan2020scalable,wang2020self}. While successful, these methods only model the event count or sequences. They do not consider the actual interaction results, like payment or purchase quantity in online shopping, and clicks/nonclicks in online advertising. Hence their problem setting is different. Despite the success of current tensor decomposition methods~\citep{Chu09ptucker,choi2014dfacto,zhe2016dintucker,zhe2015scalable,zhe2016distributed,liu2018neuralcp,pan2020streaming,tillinghast2020probabilistic,tillinghast2021nonparametric,fang2021streaming,tillinghast2021nonparametricHGP}, they all assume the factors of the entities are  static and fixed, even when the time information is incorporated. To our knowledge, our method is the first to learn these factors as trajectory functions. 
 
Our bi-level GP decomposition model can be viewed as an instance of  deep Gaussian processes (DGPs) ~\citep{damianou2013deep}. However, different from the standard DGP formulation, we conduct inverse Fourier transform over the output of the first-level GPs, before we feed them to the next-level GP. In so doing, we expect to better learn the factor trajectory functions in our problem. Our nested sparse variational inference is similar to ~\citep{salimbeni2017doubly}, where in each GP level, we also introduce a set of pseudo inputs and outputs to create sparse approximations. However, we further take advantage of our model structure to convert the (finite) GP prior in the first level as a matrix Gaussian. The benefit is that we do not need to introduce pseudo inputs in the frequency domain and so we can reduce the approximation. We use a matrix Gaussian posterior to further accelerate the computation.
\section{Experiment}
\subsection{Predictive Performance}
We evaluated \ours in three real-world benchmark datasets. (1) \textit{Beijing Air Quality}\footnote{\url{ https://archive.ics.uci.edu/ml/datasets/Beijing+Multi-Site+Air-Quality+Data}}, hourly concentration measurements of $6$ pollutants (\eg PM2.5, PM10 and SO2) in $12$ air-quality monitoring sites across Beijing from year 2013 to 2017. We thereby extracted a 2-mode (pollutant, site) tensor, including  10K measurements and the time points for different tensor entries. (2) \textit{Mobile Ads}\footnote{\url{https://www.kaggle.com/c/avazu-ctr-prediction}}, a 10-day click-through-rate dataset for mobile advertisements. We extracted a three-mode tensor \textit{(banner-position, site domain, mobile app)}, of size $7 \times 2842 \times 4127$. The tensor includes $50$K observed entry values (click number) at different time points. (3) \textit{DBLP}~\footnote{\url{https://dblp.uni-trier.de/xml/}}, the bibliographic records in the domain of computer science. We downloaded the XML database, filtered the records from year 2011 to 2021, and extracted a three-mode tensor (author, conference, keyword) of size $3731\times 1935 \times 169$ from the most  prolific authors, most popular conferences and keywords. The observed entry values are the numbers of papers published at different years. We collected $50$K entry values and their time points. 

\begin{table*}[t]
	\centering
	\small
	\begin{tabular}[c]{ccccc}
		\toprule
		\textit{Beijing Air Quality} & $R=2$ & $R=3$ & $R=5$ &  $R=7$ \\
		\hline 
		NONFAT & $\bf{0.340 \pm 0.006}$ & $ \bf{0.315	\pm 0.001}$ &	$ \bf{0.314\pm	0.001}$ &	$0.326\pm	0.006$ \\
		CPCT &	$0.997	\pm 0.001$	& $0.997	\pm 0.001$	& $1.002\pm	0.002 $&	$1.002\pm	0.002$ \\
		GPCT &	$0.372	\pm 0.001$ &	$0.366\pm	0.001$ &	 $0.363\pm	0.001$ &	$0.364\pm	0.001$ \\
		NNCT	& $0.986 \pm	0.002$ &	$0.988 \pm	0.002$ &$	0.977\pm	0.012$ &	$0.987\pm	0.003$\\
		GPDTL &	$0.884\pm	0.001$ &	$0.884	\pm 0.001$	& $0.885\pm	0.001$ &	 $0.884 \pm	0.001$ \\
		NNDTL & $	0.356 \pm	0.003$ &	$0.358 \pm	0.005$ &	$0.333\pm	0.003$ &	$\bf{0.315\pm	0.002}$\\
		GPDTN & $	0.884 \pm	0.001 $&	$0.884 \pm	0.001$ &	$0.884 \pm	0.001$ & $	0.884 \pm	0.001$ \\
		NNDTN	& $0.365 \pm	0.005$ &	$0.337 \pm	0.006$ &	$0.336 \pm	0.003$ &	${0.319 \pm	0.005}$\\
		\midrule
		\textit{Mobile Ads}    &           \\ 
		\midrule
		NONFAT &	$0.652 \pm	0.002$ &	$\bf{0.635 \pm	0.003}$ &$\bf{0.638 \pm	0.006}$ &	$\bf{0.637 \pm	0.005}$ \\
		CPCT &	$1.001 \pm	0.004$ &	$0.986 \pm	0.018$ &$	1.009 \pm	0.009$ &	$0.971 \pm	0.010$ \\
		GPCT	&$ 0.660 \pm	0.003$ &	$0.661 \pm	0.003$ &	$0.662 \pm	0.001$ &$	0.659 \pm	0.003$ \\
		NNCT &	$0.822 \pm	0.001$ &	$0.822 \pm	0.001$ &	$0.822 \pm	0.001$ &	$0.822\pm	0.001$ \\
		GPDTL &	$0.714 \pm	0.006$ &	$0.695 \pm	0.004$ &	$0.695 \pm	0.004$ &	$0.695 \pm	0.003$ \\
		NNDTL &	 $\bf{0.646	 \pm 0.003}$ &	$0.646 \pm	0.002$ &$	0.642 \pm	0.003 $&	$0.640 \pm	0.003$\\
		GPDTN & $	0.667 \pm	0.003$	&$ 0.661 \pm	0.003$ &$	0.668 \pm	0.003$ &	$0.669 \pm	0.003$ \\
		NNDTN &	$0.646 \pm	0.004$ &	$0.645 \pm	0.002$ &$	0.640 \pm	0.003$ &	$0.638 \pm	0.003$\\
		\midrule
		\textit{DBLP}    &           \\ 
		\midrule
		NONFAT &	$\bf{0.188\pm 0.003} $& 	$ \bf{0.188\pm	0.003}$ &	$0.189	\pm 0.003$&	$\bf{0.189\pm	0.003}$ \\
		CPCT &	$1.004\pm	0.003$ &	$1.004 \pm	0.002$ & $	1.005 \pm	0.002$& $	1.001 \pm	0.004$ \\
		GPCT & $	0.189 \pm	0.003 $& $	0.191 \pm	0.003 $& $	0.192 \pm	0.003 $& $	0.196 \pm	0.003 $ \\
		NNCT & $	0.188	\pm 0.003$& $	{0.188 \pm	0.003}$& $\bf{	0.188\pm	0.003}$& $	0.189 \pm	0.003$ \\
		GPDTL	& $ 0.208 \pm	0.004$& $	0.223 \pm	0.003$& $	0.221\pm	0.003$& $	0.224\pm	0.003$ \\
		NNDTL & $	0.188 \pm	0.003$& $	0.188\pm	0.003$& $	0.189\pm	0.003$& $	0.189\pm	0.003$ \\
		GPDTN	& $0.206 \pm	0.002	$& $0.218 \pm	0.003$& $	0.224 \pm	0.003$& $	0.225 \pm	0.002$ \\
		NNDTN	& $0.188\pm	0.003	$& $0.188	\pm 0.003$& $	0.188\pm	0.003$& $	0.189\pm	0.003$ \\
		\bottomrule
	\end{tabular}
	\caption{\small Root Mean-Square Error (RMSE). The results were averaged over five runs.  }
	\label{tb:rmse}
\end{table*}
\begin{table*}[t]
	\centering
	\small
	\begin{tabular}[c]{ccccc}
		\toprule
		\textit{Beijing Air Quality} & $R=2$ & $R=3$ & $R=5$ &  $R=7$ \\
		\hline 
		NONFAT &	$\bf{-0.343 \pm	0.018} $&$	\bf{-0.264 \pm	0.003}$&$	\bf{-0.260\pm	0.004}$&$	\bf{-0.297\pm	0.017}$ \\			
		GPCT & $	-0.420 \pm	0.001$&$	-0.406 \pm	0.001$&$	-0.401 \pm	0.001$&$	-0.401 \pm	0.001$ \\
		GPDTL	& $-1.299 \pm	0.001$&$	-1.299 \pm	0.001$&$	-1.299 \pm	0.001$&$	-1.299\pm	0.001$ \\							
		GPDTN & $	-1.299\pm	0.001$&$	-1.299	\pm 0.001$&$ 	-1.299\pm	0.001$&$	-1.299\pm	0.001$ \\
		\midrule
		\textit{Mobile Ads}    &           \\ 
		\midrule
		NONFAT & $	\bf{-0.726 \pm	0.004} $&$	\bf{-0.705 \pm	0.003}$&$\bf{	-0.709\pm	0.007}$&$	\bf{-0.706 \pm	0.008} $ \\			
		GPCT & $	-0.733 \pm	0.002$&$	-0.737 \pm	0.005$&$	-0.734\pm	0.004$&$	-0.735 \pm	0.004$ \\
		GPDTL &$	-1.843 \pm	0.009$&$	-1.807 \pm	0.006$&$	-1.822\pm	0.008$&$	-1.830\pm	0.003$ \\
		GPDTN	 & $-0.774 \pm	0.003$&$	-0.762\pm	0.004$&$	-0.804\pm	0.006$&$	-0.806\pm	0.003$ \\
		\midrule
		\textit{DBLP}    &           \\ 
		\midrule
		NONFAT & $	\bf{0.201 \pm	0.019} $&$	\bf{0.201\pm	0.019}$&$	\bf{0.199 \pm	0.017}$&$	\bf{0.199	\pm 0.017}$ \\
		GPCT & $	{0.129 \pm	0.009} $&$	{0.105	\pm 0.009} $&${	0.104 \pm	0.011}$&$ {0.087	\pm 0.013}$ \\
		GPDTL & $ 0.102	\pm 0.023	$&$0.004 \pm	0.025$&$	0.035\pm	0.019$&$	0.022\pm	0.019$ \\						
		GPDTN &	$0.114 \pm	0.012$&$	0.041 \pm	0.019$&$	0.019 \pm	0.020$&$	0.013	\pm 0.015$ \\
		\bottomrule
	\end{tabular}
	\caption{Test log-likelihood. The results were averaged from five runs.}
	\label{tb:ll}
	\vspace{-0.2in}
\end{table*}

We compared with the following popular and/or state-of-the-art methods for dynamic tensor decomposition.  (1) CPCT~\citep{zhang2021dynamic}, the most recent continuous-time decomposition algorithm, which uses polynomial splines to estimate the coefficients in the CP framework as a temporal function (see $\blambda$ in \eqref{eq:cp}). (2) GPCT,  continuous-time GP decomposition, which extends~\citep{xu2012infinite,zhe2016distributed} by placing the time $t$ in the GP kernel to learn the entry value as a function of both the latent factors and time $t$, \ie $m_\bell = f(\u^1_{\ell_1}, \ldots, \u^K_{\ell_K},t)$. (3) NNCT, continuous-time neural network decomposition, which is similar to~\citep{costco} but uses time $t$ as an additional input to the NN decomposition model. (4) GPDTL, discrete-time GP decomposition with linear dynamics, which expands the tensor with a time mode and jointly estimates the time factors and other factors with GP decomposition~\citep{zhe2016distributed}. In addition, we follow~\citep{xiong2010temporal} to introduce a conditional prior over consecutive steps, $p(\t_{j+1}|\t_j) = \N(\t_{j+1}|\C \t_j + \b, v\I)$. Note this linear dynamic model is more general than the one in~\citep{xiong2010temporal} since the latter corresponds to $\C=\I$ and $\b = \0$. (5) GPDTN, discrete-time GP decomposition with nonlinear dynamics. It is similar to GPDTL except that the prior over the time factors becomes $p(\t_{j+1}|\t_j) = \N(\t_{j+1}|\sigma(\C \t_j) + \b, v\I)$ where $\sigma(\cdot)$ is the nonlinear activation function. Hence, this can be viewed as an RNN-type transition. (6) NNDTL, discrete-time NN decomposition with linear dynamics, similar to GPDTL but using NN decomposition model. (7) NNDTN, discrete-time NN decomposition with nonlinear dynamics, which is similar to GPDTN that employs RNN dynamics over the time steps. 

{\bf Experiment Setting.} We implemented all the methods with PyTorch~\citep{paszke2019pytorch}. For all the GP  baselines, we used SE kernel and followed~\citep{zhe2016dintucker} to use sparse variational GP framework~\citep{hensman2013gaussian} for scalable posterior inference, with $100$ pseudo inputs. For \ours, the number of pseudo inputs for both levels of GPs was set to $100$.  For NN baselines, we used three-layer neural networks, with $50$ neurons per layer and \texttt{tanh} as the activation. For nonlinear dynamic methods, including GPDTN and NNDTN, we used \texttt{tanh} as the activation.  For CPCT, we used $100$ knots for polynomial splines. For discrete-time methods, we used 50 steps. We did not find improvement with more steps. All the models were trained with stochastic mini-batch optimization. We used ADAM~\citep{adam} for all the methods, and the mini-batch size is $100$. The learning rate was chosen from $\{10^{-4}, 5 \times 10^{-4}, 10^{-3}, 5 \times 10^{-3}, 10^{-2}\}$. To ensure convergence, we ran each methods for 10K epochs. To get rid of the fluctuation of the error due to the stochastic model updates, we computed the test error after each epoch and used the smallest one as the result.  We varied the number of latent factors $R$ from \{2, 3, 5, 7\}. For \ours, $R$ is the number of factor trajectories. We followed the standard testing procedure as in~\citep{xu2012infinite,kang2012gigatensor,zhe2016distributed} to randomly sample $80\%$ observed tensor entries (and their timestamps) for training and tested the prediction accuracy on the remaining entries. We repeated the experiments for five times and calculated the average root mean-square-error (RMSE) and its standard deviation. 

\textbf{Results.} As we can see from Table \ref{tb:rmse}, in most cases, our methods \ours outperforms all the competing approaches, often by a large margin. In addition, \ours  always achieves better prediction accuracy than GP methods, including GPCT, GPDTL and GPDTN. In most cases, the improvement is significant ($p<0.05$). It shows that our bi-level GP decomposition model can indeed improves upon the single level GP models.  Only in a few cases, the prediction error of \ours is slightly worse an NN approach. Note that, \ours learns a trajectory for each factor, and it is much more challenging than learning fixed-value factors, as done by the competing approaches.

Furthermore, we examined our method in a probabilistic sense. We reported the test log-likelihood of \ours and the other probabilistic approaches, including GPCT, GPDTL and GPDTN, in  Table \ref{tb:ll}. We can see that \ours largely improved upon the competing methods in all the cases, showing that \ours is advantageous not only in prediction accuracy but also in uncertainty quantification, which can be important in many decision tasks, especially with sparse and noisy data. 

\begin{figure}
	\centering
	\setlength{\tabcolsep}{0pt}
	\begin{tabular}[c]{ccc}
		\begin{subfigure}[b]{0.17\textwidth}
			\centering
			\includegraphics[width=\linewidth]{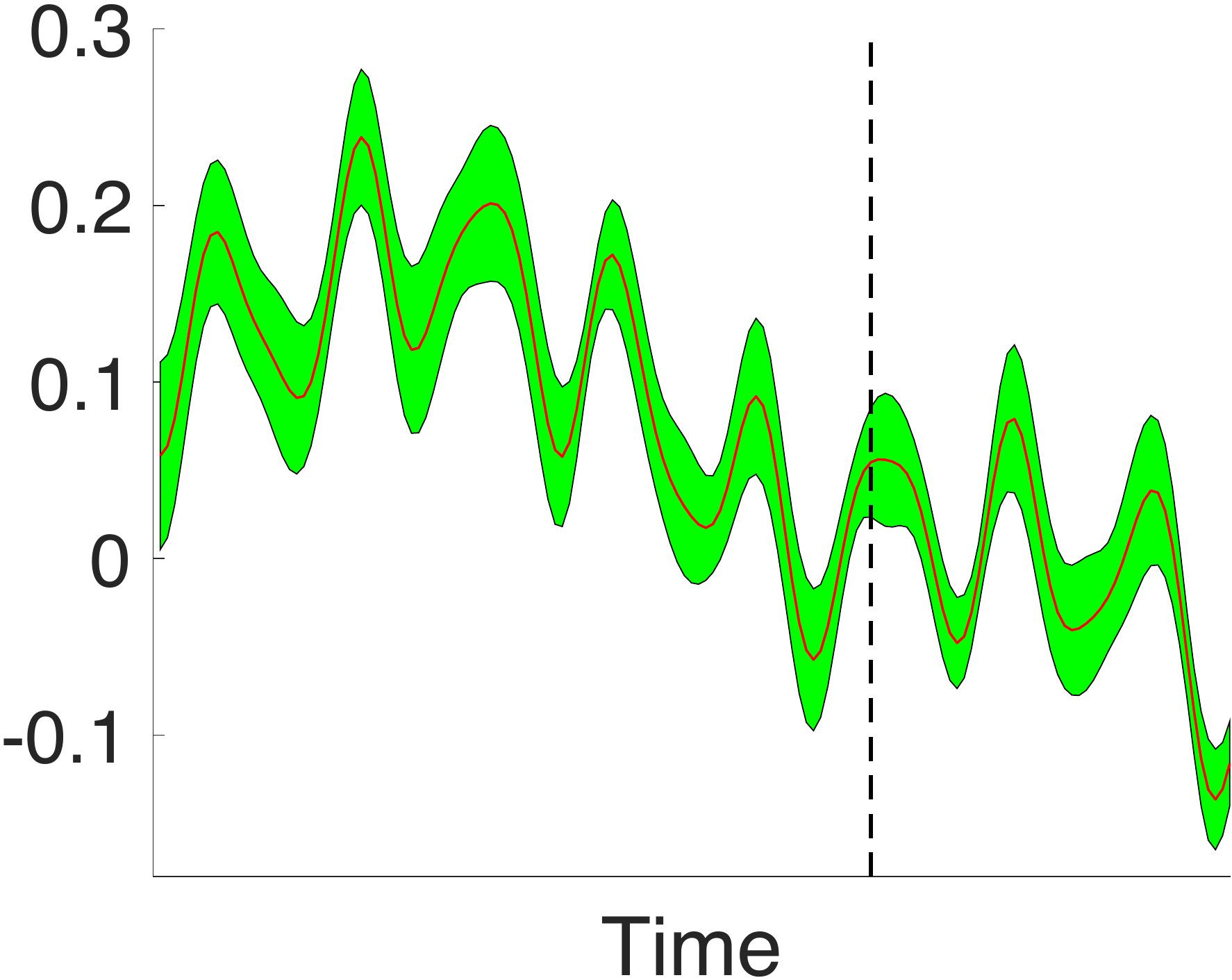}
			\caption{$u^1_{1,1}(t)$}
		\end{subfigure} &
		\begin{subfigure}[b]{0.17\textwidth}
			\centering
			\includegraphics[width=\linewidth]{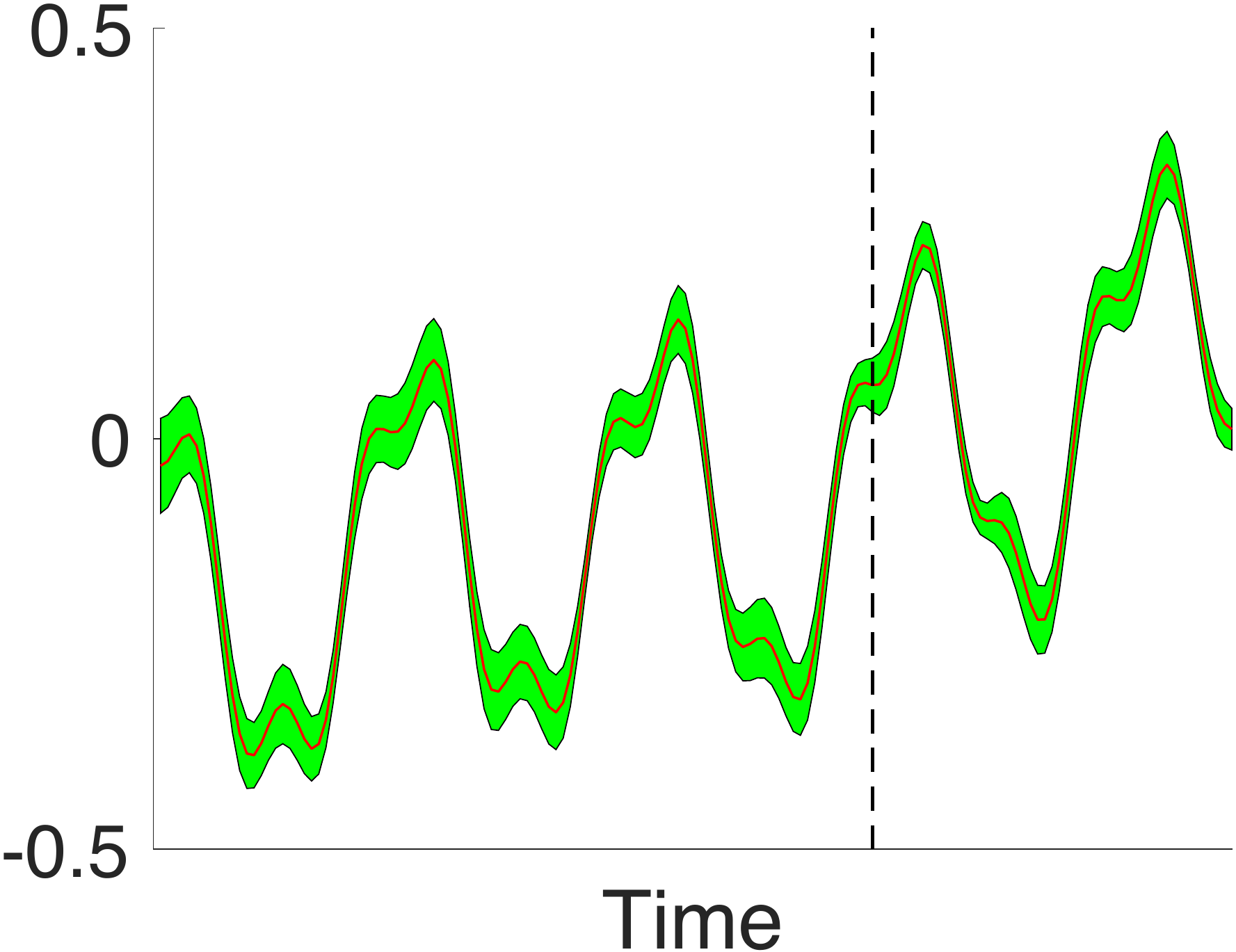}
			\caption{$u^1_{1,2}(t)$} 
		\end{subfigure} &
		\begin{subfigure}[b]{0.17\textwidth}
			\centering
			\includegraphics[width=\linewidth]{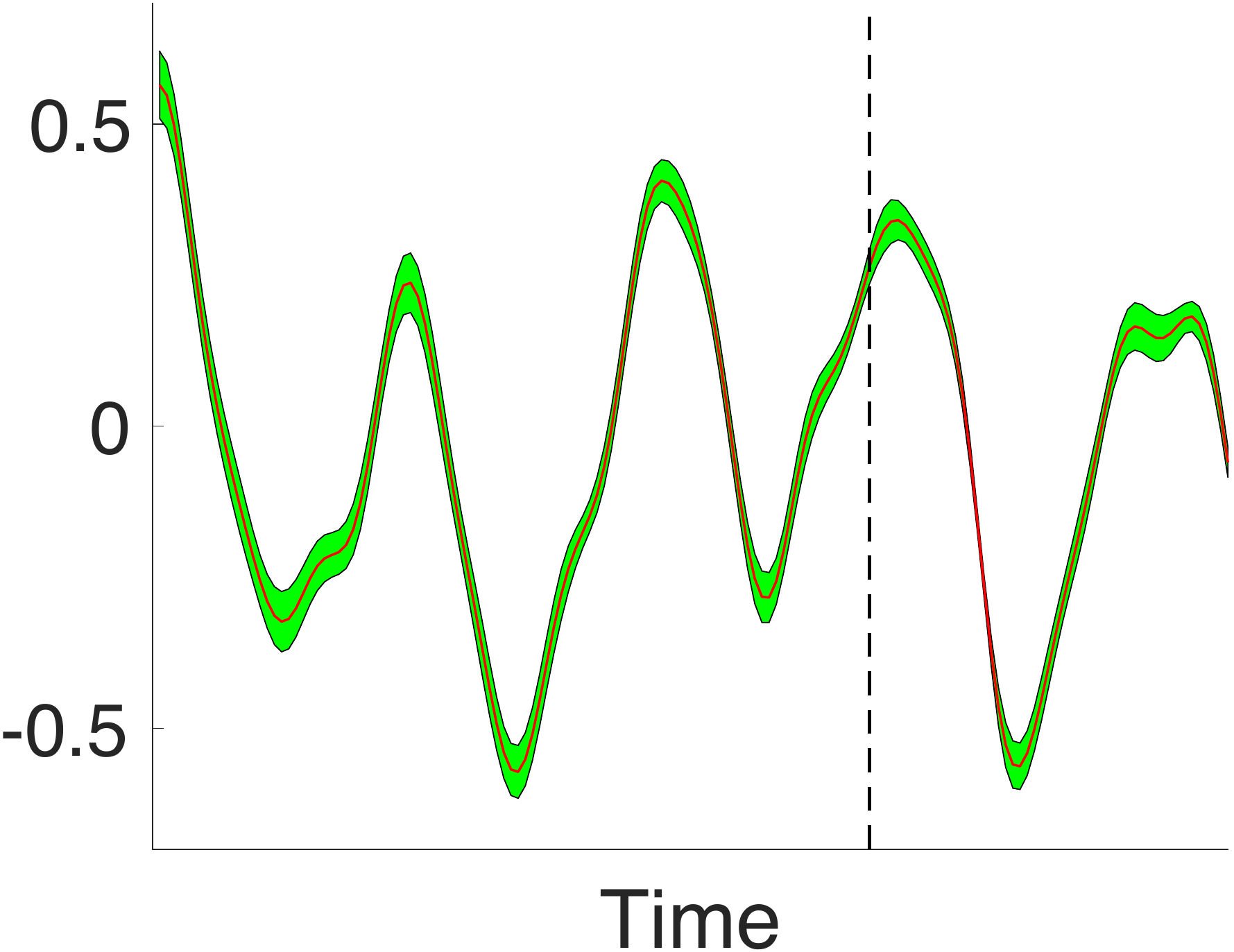}
			\caption{$u^1_{1,3}(t)$} 
		\end{subfigure}\\
		\begin{subfigure}[b]{0.17\textwidth}
		\centering
		\includegraphics[width=\linewidth]{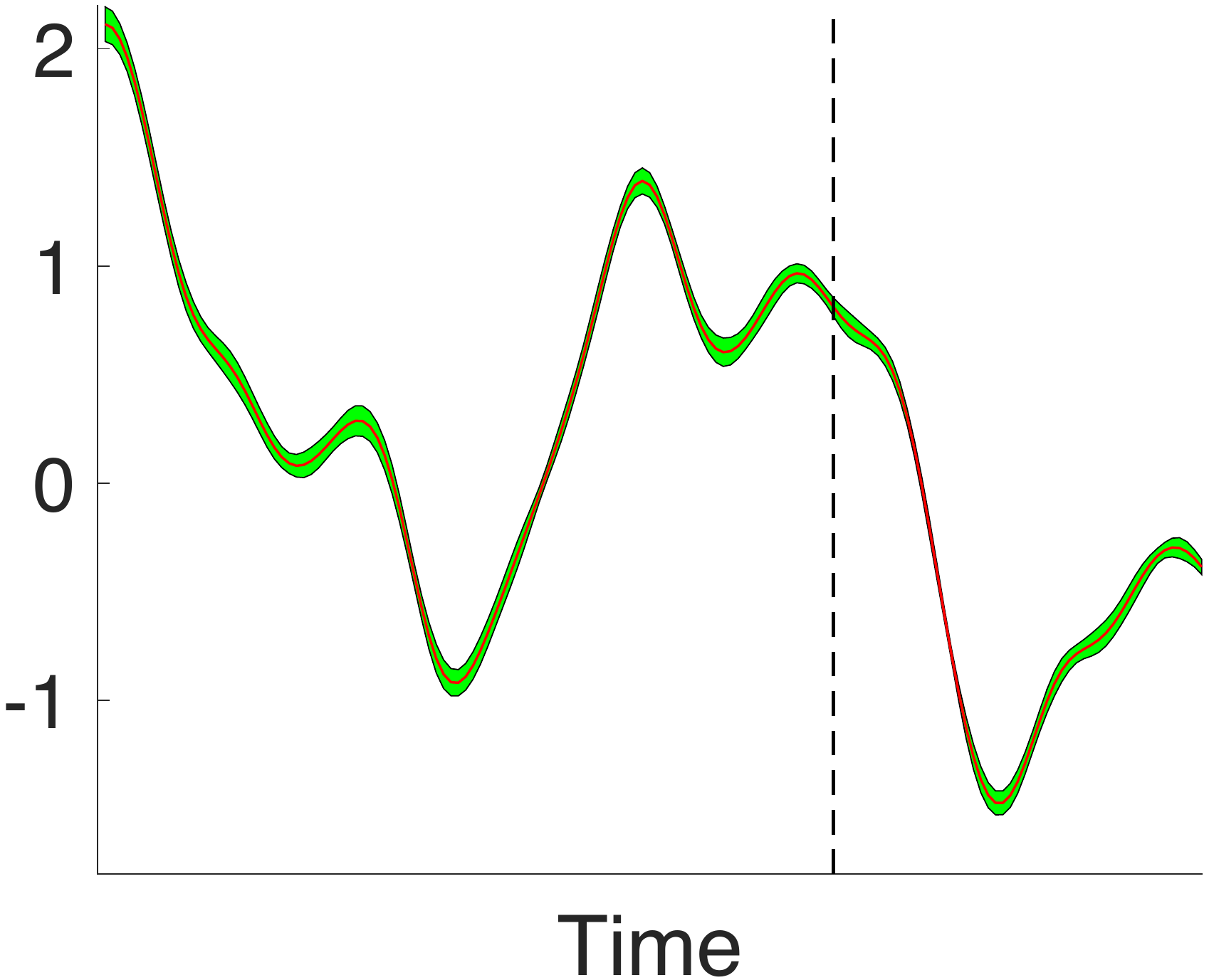}
		\caption{$u^2_{2,1}(t)$}
	\end{subfigure} &
	\begin{subfigure}[b]{0.17\textwidth}
		\centering
		\includegraphics[width=\linewidth]{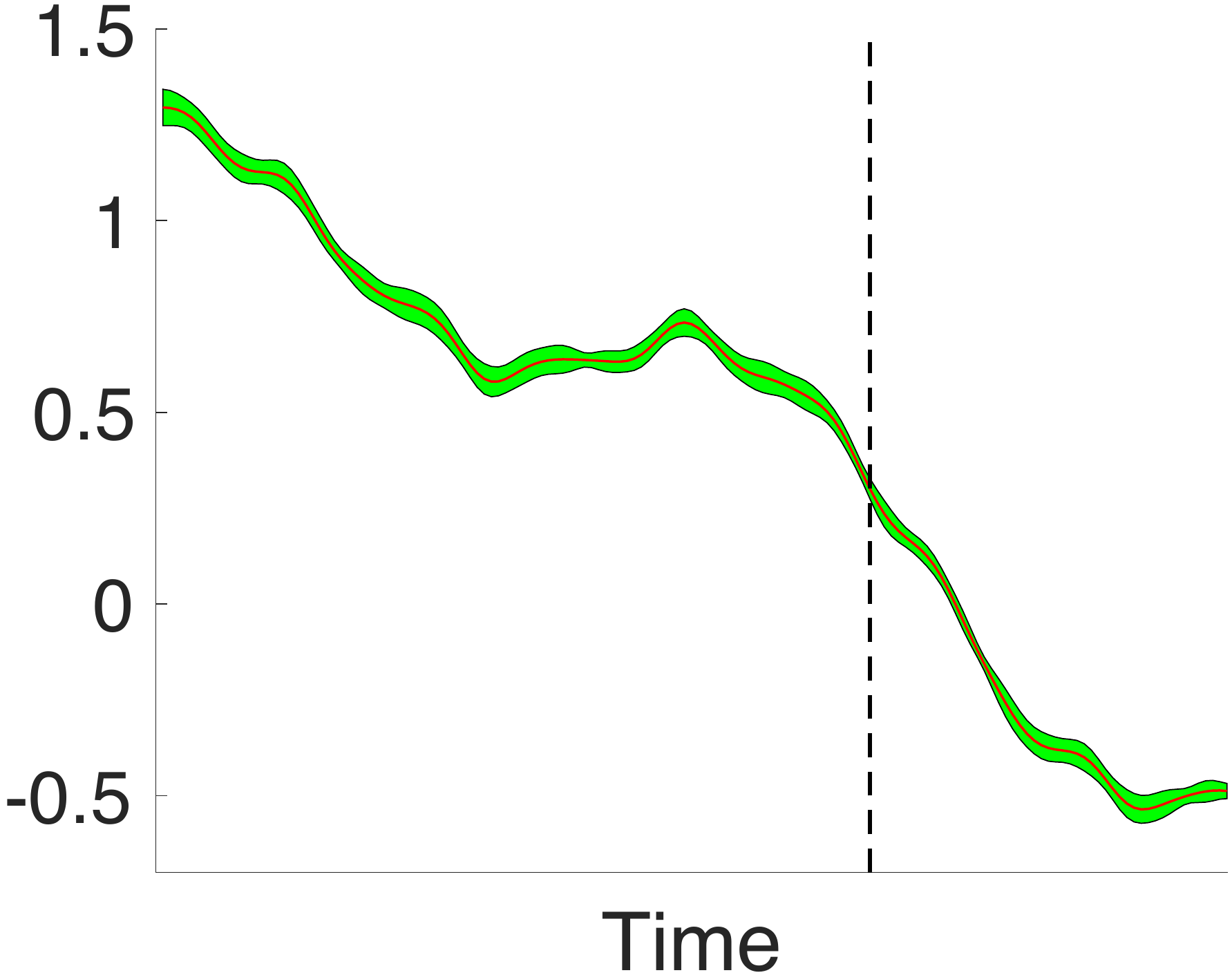}
		\caption{$u^2_{2,2}(t)$} 
	\end{subfigure} &
	\begin{subfigure}[b]{0.17\textwidth}
		\centering
		\includegraphics[width=\linewidth]{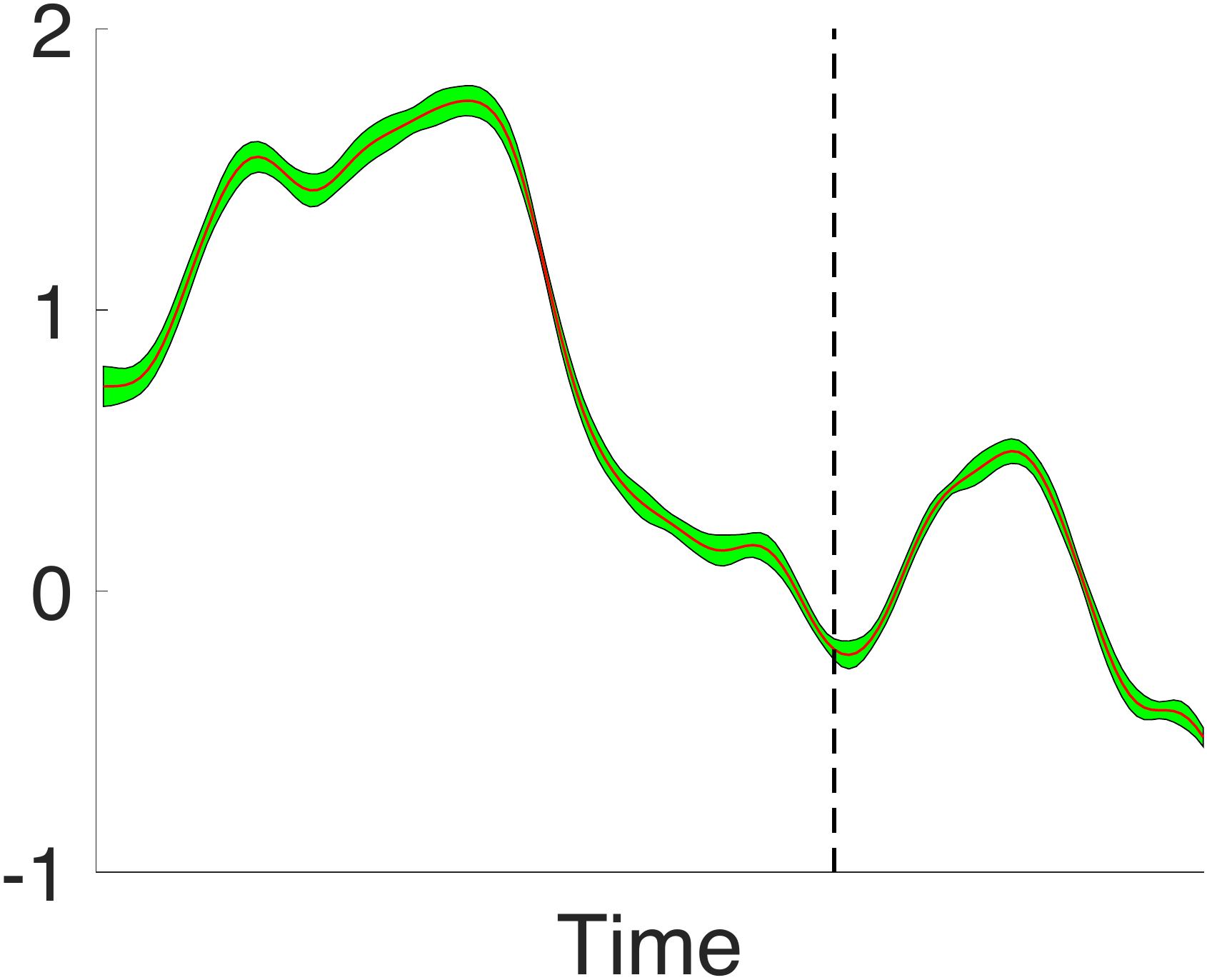}
		\caption{$u^2_{2,3}(t)$} 
	\end{subfigure}
	\end{tabular}
	\caption{\small The learned factor trajectories. }
	\label{fig:trajectory}
\end{figure}
\begin{figure}
	\centering
	\setlength{\tabcolsep}{0pt}
	\begin{tabular}[c]{cc}
		\begin{subfigure}[b]{0.25\textwidth}
			\centering
			\includegraphics[width=\linewidth]{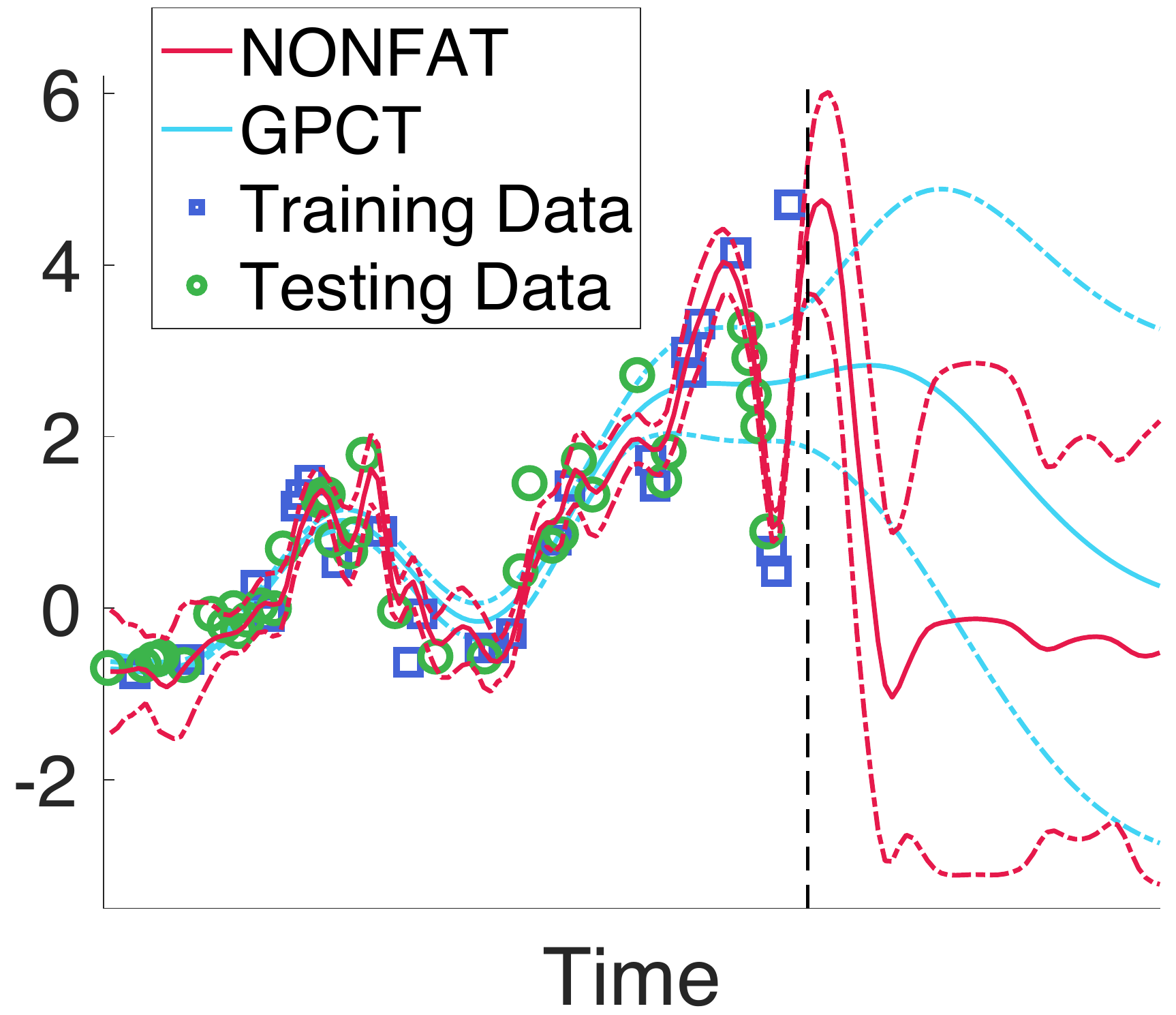}
			\caption{(6, 6)}
		\end{subfigure} &
		\begin{subfigure}[b]{0.25\textwidth}
			\centering
			\includegraphics[width=\linewidth]{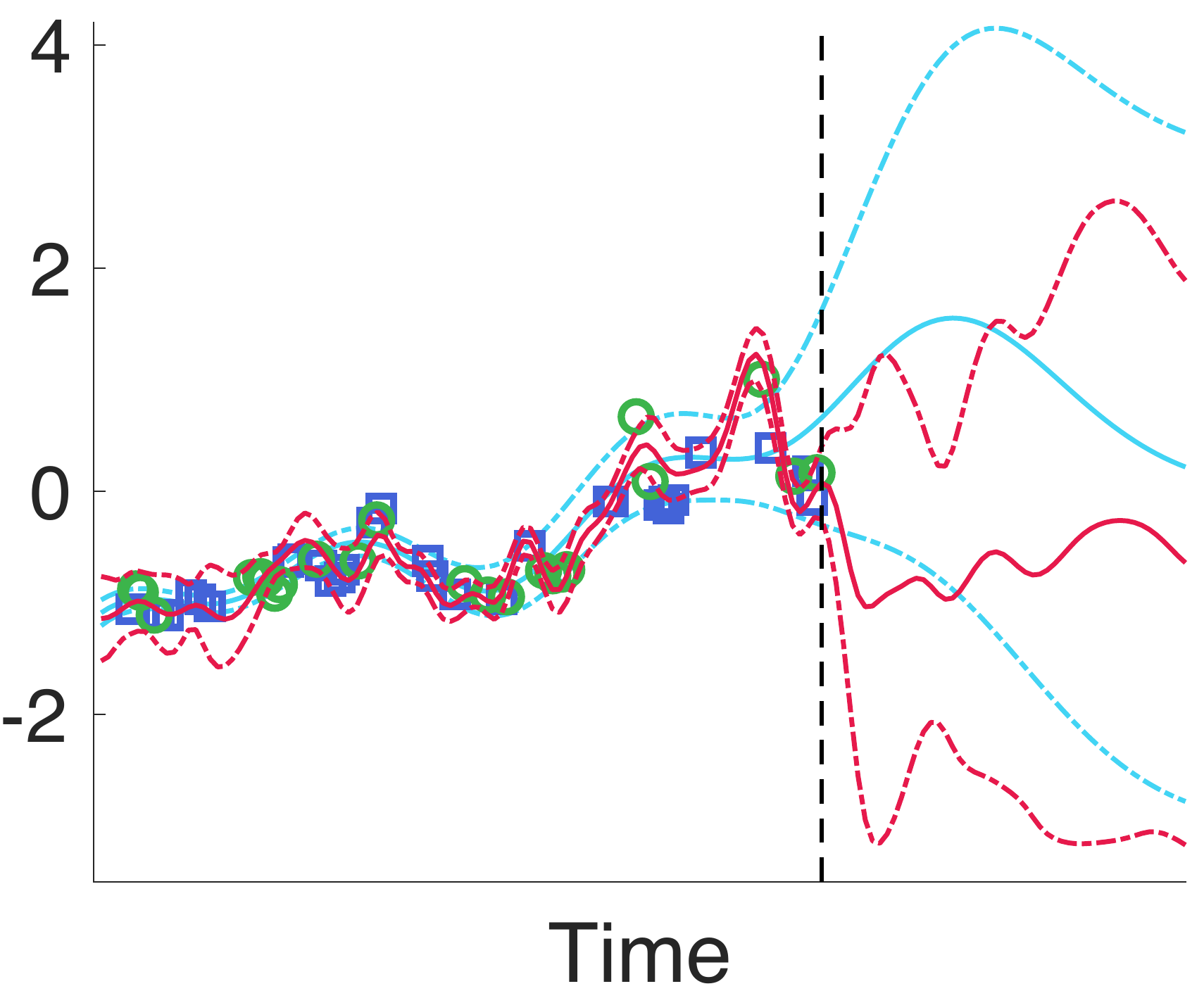}
			\caption{(7,4)} 
		\end{subfigure} 
	\end{tabular}
	\caption{\small Entry value prediction.}
	\label{fig:pred-curve}
\end{figure}
\subsection{Investigation of Learning Result }
Next, we investigated if our learned factor trajectories exhibit patterns and how they influence the prediction. To this end, we set $R=3$ and ran \ours on \textit{Beijing Air Quality} dataset. We show the learned factor trajectories for the first monitoring site in mode 1 in Fig. \ref{fig:trajectory} a-c,  and second pollutant (SO2) in mode 2 in Fig. \ref{fig:trajectory} d-f. As we can see, they show different patterns. First, it is interesting to see that all the trajectories for the site exhibit periodicity, but with different perturbation, amplitude, period, \etc This might relate to the working cycles of the sensors in the site.  The trajectories for the pollutant is much less periodic and  varies quite differently. For example, $u^2_{2, 1}(t)$ decreases first and then increases, while $u^2_{2, 3}(t)$ increases first and then decreases, and $u^2_{2, 2}(t)$ keeps the decreasing trend. They represent different time-varying components of the pollutant concentration. Second, the vertical dashed line is the boundary of the training region. We can see that all the trajectories extrapolate well. Their posterior mean and standard deviation are stable outside of the training region (as stable as inside the training region). It demonstrates our learning from the frequency domain can yield robust trajectory estimates. 

Finally, we showcase the prediction curves of two entries in Fig. \ref{fig:pred-curve}. The prediction made by our factor trajectories can better predict the test points, as compared with GPCT. More important, outside the training region, our predictive uncertainty is much smaller than GPCT (right to the dashed vertical line), while inside the training region, our predictive uncertainty is not so small as GPCT that is close to zero. This  shows \ours gives more reasonable uncertainty quantification in both interpolation and extrapolation.

\section{Conclusion}
We have presented NONFAT, a novel nonparametric Bayesian method to learn factor trajectories for dynamic tensor decomposition. The predictive accuracy of \ours in real-world applications is encouraging and the learned trajectories show interesting temporal patterns. In the future, we will investigate the meaning of these patterns more in-depth and apply our approach in more applications. 

\section*{Acknowledgments}
This work has been supported by NSF IIS-1910983 and NSF CAREER Award IIS-2046295.

\bibliographystyle{apalike}
\bibliography{EmbedTraj}



\end{document}